\documentclass[11pt,a4paper]{article}
\usepackage[hyperref]{emnlp2017}
\usepackage{times}
\usepackage{latexsym}
\usepackage{url}
\usepackage{amsmath}
\usepackage{graphicx}
\usepackage{multirow}
\usepackage{bm}
\usepackage{CJKutf8}
\emnlpfinalcopy

\title{Assigning Personality/Identity to a Chatting Machine \\ for Coherent Conversation Generation}
\author{Qiao Qian$^1$, Minlie Huang$^1$, Haizhou Zhao$^2$, Jingfang Xu$^2$, Xiaoyan Zhu$^1$\\
$^1$State Key Laboratory of Intelligent Technology and Systems \\
Tsinghua National Laboratory for Information Science and Technology\\
Dept. of Computer Science and Technology, Tsinghua University, Beijing 100084, PR China \\
$^2$Sogou, Inc. , Beijing, 100084, PR China \\
{\tt qianqiaodecember29@126.com}, \quad {\tt aihuang@tsinghua.edu.cn}, \\
{\tt zhaohaizhou@sogou-inc.com}, \quad {\tt xujingfang@sogou-inc.com}, \\
{\tt zxy-dcs@tsinghua.edu.cn}}

\date{}

\begin{document}
\begin{CJK*}{UTF8}{gbsn}
\maketitle
\begin{abstract}
Endowing a chatbot with personality or an identity is quite challenging but critical to deliver more realistic and natural conversations. 
In this paper, we address the issue of generating responses that are coherent to a pre-specified agent profile. 
We design a model consisting of three modules: a profile detector to decide whether a post should be responded using the profile and which key should be addressed, a bidirectional decoder to generate responses forward and backward starting from a selected profile value, and a position detector that predicts a word position from which decoding should start given a selected profile value. We show that general conversation data from social media can be used to generate profile-coherent responses.
Manual and automatic evaluation shows that our model can deliver more coherent, natural, and diversified responses.

\end{abstract}

\section{Introduction}
Generating human-level conversations by machine has been a long-term goal of AI since the Turing Test \cite{turing1950computing}. However, as argued by \cite{vinyals2015neuralConversal}, the current conversation generation models are still unable to deliver realistic conversations to pass the Test. Amongst the many limitations, the lack of a coherent personality is one of the most challenging difficulties. Though personality is a well-defined concept in psychology \cite{norman1963toward,gosling2003very}, while in this paper, the personality of a chatbot refers to the character that the bot plays or performs during conversational interactions. In this scenario, 
personality can be viewed as a composite of the identity (the background and profile) that a chatbot is endowed with, linguistic style that an agent exhibits during interactions \cite{Walker97improvinglinguistic}, and many more explicit and implicit cues that may portray character. Though personality is a more abstract and broad concept, we use {\it personality, profile, and identity} interchangeably in this paper. 

\begin{table} [!htp]
\centering
	\begin{tabular}{|l|}
		\hline
		\textbf{General seq2seq model} \\
		\hline
		User: Are you a boy or a girl?	\\ 
		Chatbot: I am a boy. \\ 
		User: Are you a girl? \\ 
		Chatbot: Yes, I am a girl. \\ 
		\hline 
		\textbf{Our model with personality} \\
		\hline
		User: Are you a boy or a girl? \\ 
		Chatbot: I am a handsome boy. \\ 
		User: Are you a girl? \\ 
		Chatbot: No, I am a boy. 	\\
		\hline 
	\end{tabular} 
	\caption{Exemplar conversations with/without coherent personality.}
	\label{tab-examplar-conv}
\end{table}

\begin{table} [!htp]
\vspace{1mm}
\centerline{
	\begin{tabular}{|l|l|}
		\hline
		\textbf{Profile key} & \textbf{Profile value} \\
		\hline
		Name & 汪仔 (Wang Zai) \\
		\hline
		Age   & 三岁 (3) \\
		\hline
		Gender & 男孩 (Boy) \\
		\hline
		Hobbies & 动漫 (Cartoon) \\
		\hline
		Speciality & 钢琴 (Piano) \\
		\hline 
	\end{tabular}}
	\caption{An exemplar agent profile. }
	\label{tab-profile}
\end{table}

Endowing a chatbot with personality is well motivated by the simple example as shown in Table \ref{tab-examplar-conv}. We can clearly see that a general sequence-to-sequence (Seq2Seq) model cannot exhibit coherent personality/identity while our model is more \textbf{coherent to a given identity}. 
The motivation is also verified by \cite{yu2016strategy} which reports that users ask for much personal information of a chatbot in human-machine interaction. It is evident that personal information of a chatbot is much attended by users during conversation, particularly at the early stage of interaction.
 
The recent work dealing with personality in large-scale conversation generation can be seen in \cite{li2016Persona} where speaker-specific conversation style is learned by user embedding.
Another work which models user personalization can be seen in \cite{al2016conversational}, with a similar technique of user embedding.
Both studies require dialogue data from each user to model her/his personality.
 
The major departure to the previous works lies in: \textbf{First},
we address the problem of endowing a chatbot with a given identity. 
Such a task requires chatbots to {\it generate not only consistent responses, but also responses that are coherent to its pre-specified identity/personality}.   
\textbf{Second}, the previous works on personality modeling require conversation data from each user, however, it's impractical here since dialogue data from the chatbot are unavailable before the release of the chatbot. Instead of just learning personality from dialogue data, our work can assign a desired identity to a chatbot by making use of general conversation data from social media.  
%
%
%
%
Our contributions lie in two folds:
\begin{itemize}

\item We investigate the problem of endowing a chatbot with a given identity and enabling a chatbot to generate responses that are coherent to its given identity. Instead of learning personality from dialogue data, our work can assign a desired identity to a chatbot. 

\item To address the problem, we propose a model consisting of a profile detector, a position detector, and a bidirectional decoder. Post-level and session-level evaluation shows that when giving an agent profile, our model can generate more coherent responses with more language variety.

\end{itemize}

\section{Related Work}
There has been a large amount of work for dialogue/conversation generation. These works can be categorized into task-oriented \cite{young2013pomdp,wen2016network,bordes2016learning} or chat-based. Recently, researchers found that social data such as Twitter/Weibo posts and 
replies~\cite{ritter2011dataSMT,shang2015NeuralRespondingMachine}, and movie dialogues  can be used to learn and generate spoken language.

Large-scale conversation generation with social media data was firstly proposed in \cite{ritter2011dataSMT} and has been greatly advanced by  applying sequence-to-sequence models  \cite{sutskever2014seq2seq,vinyals2015neuralConversal,sordoni2015neuralContextSensitive,shang2015NeuralRespondingMachine,SerbanSBCP15,serban2016HRED}. 
Many studies are focusing on improving the generation quality. 
These works include: dealing with unknown words \cite{gu2016CopySeq2Seq,gulcehre2016pointingUNK}, avoiding universal responses \cite{li2016diversityMMI}, generating more diverse and meaningful responses \cite{xing2017topicAware,mou2016sequenceBackwardForward}, and many more.

As argued by \cite{vinyals2015neuralConversal}, it's still quite impossible for current chatbots 
to pass the Turing Test, while one of the reasons is the lack of a coherent personality. Though personality has been well defined in psychology \cite{norman1963toward}, it is implicit, subtle, and challenging to formally define in statistical language generation.
Linguistic style can be an indicator of personality \cite{mairesse2006automatic,mairesse2007using}, and conversation can be clues for personality recognition \cite{Walker97improvinglinguistic,walker2012annotated}. In reverse, spoken language can be generated in accordance to particular personality \cite{mairesse2007personage}. 

A first attempt to model persona can be seen in \cite{li2016Persona} where the authors proposed to learn speaker-specific conversation style by user embedding. 
Our work differs from this work significantly: our task is to endow the chatbot with a fixed personality while \cite{li2016Persona} learns personalized conversational styles. In other words, our task requires to generate not only consistent responses, but also responses that are coherent to the chatbot's prespecified identity. 
Further, \cite{li2016Persona} requires many dialogue data from each user while our work has no such requirement.

Another related work is generative question answering (GenQA) \cite{yin2015GenQA} which generates a response containing an answer extracted from a knowledge base (KB). However, endowing a chatbot with personality is more than just question answering over KB, where there arise challenging problems such as semantic reasoning and conversation style modeling. Further, GenQA requires that the answer from KB must appear in the response to provide sufficient supervision while our work avoids the limitation by applying a position detector during training.

\section{Model}
\subsection{Task Definition}
The task can be formally defined as follows:
given a post $\bm{x}=x_1x_2\cdots x_n$, and an agent profile defined as a set of key-value pairs \{$<k_i,v_i>|i=1,2,\cdots,K$\}, the task aims to generate 
a response $\bm{y}=y_1y_2\cdots y_m$ that is coherent to the agent profile.
The generation process can be briefly stated as below:

\begin{equation}
    \label{eq:overview}
    \begin{split}
        & \bm{P}(\bm{y}|\bm{x},\{<k_i,v_i>\}) \\
        = & \bm{P}(z=0|\bm{x}) \cdot \bm{P}^{fr}(\bm{y}|\bm{x}) \\
        + & \bm{P}(z=1|\bm{x}) \cdot \bm{P}^{bi}(\bm{y}|\bm{x},\{<k_i,v_i>\})
    \end{split}
\end{equation}
where $\bm{P}(z|\bm{x})$ is the probability of using the agent profile given post $\bm{x}$, which is computed by the \textit{Profile Detector};
$\bm{P}^{fr}(\bm{y}|\bm{x})=
\prod_{t=1}^m \bm{P}^{fr}(y_t|y_{<t},\bm{x}) $ is given by a general forward decoder, the same as \cite{sutskever2014seq2seq}, and $\bm{P}^{bi}(\bm{y}|\bm{x},\{<k_i,v_i>\})$ 
is given by a \textit{Bidirectional Decoder} which will be described later.

Note that the post/response pairs $<\bm{x},\bm{y}>$ are collected from social media, and the agent profile value may not occur
in the response $\bm{y}$. This leads to the discrepancy between training and test, which will be addressed in the \textit{Position Detector} section.

\subsection{Overview}

\begin{figure}[!htp]
	\centering
	\includegraphics[width=0.50\textwidth]{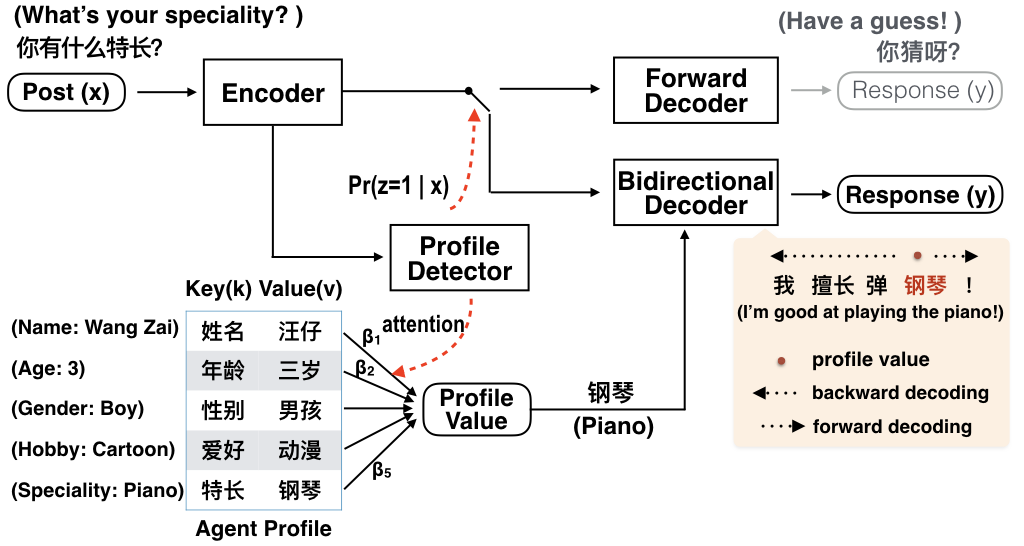}
	\caption{The overall process. 
	} \label{fig_overview}
\end{figure}

\begin{figure}[!htp]
	\centering
	\includegraphics[width=0.50\textwidth]{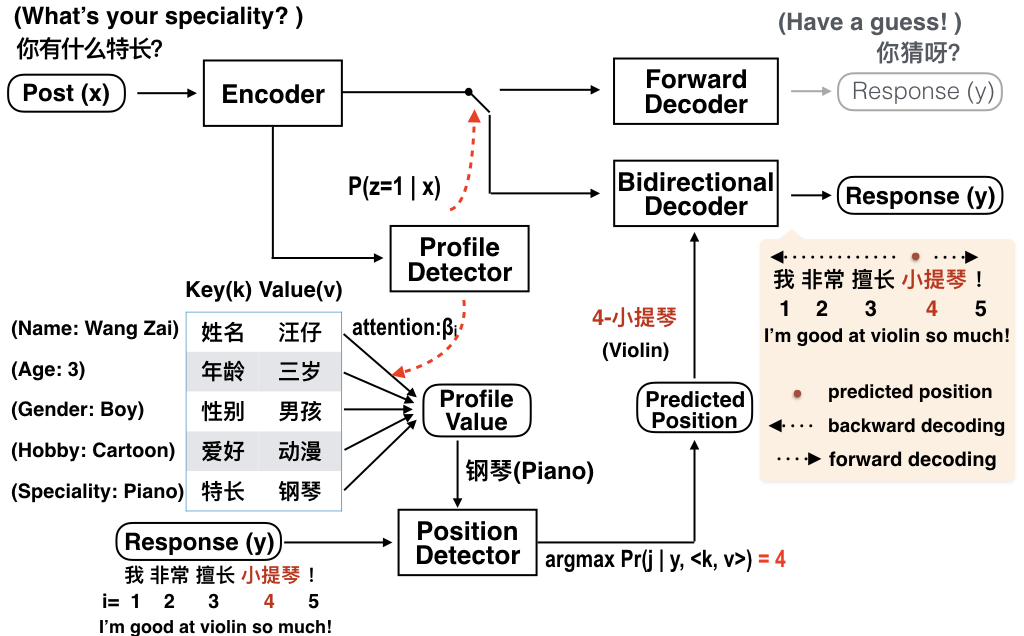}
	\caption{The training process of the model. Given a pair $<\bm{x},\bm{y}>$, the position detector will predict a position 小提琴-4(violin) at which the profile value 钢琴(piano) can be replaced, and the position will be used to train the bidirectional decoder. } \label{fig_overview_training}
\end{figure}

Our model works as follows (see Figure \ref{fig_overview}): given a post, the profile detector will predict whether the agent profile should be used.
If not, a general seq2seq decoder will be used to generate the response; otherwise, the profile detector will further select an appropriate profile key and its value.
Starting from the selected profile value, a response will be generated forward and backward by the bidirectional decoder. To better train the bidirectional decoder (see Figure \ref{fig_overview_training}), the position detector addresses the discrepancy issue between training and test, by predicting a word position from which decoding should start given the selected profile value. Note that the position detector will not participate in generation during test.

%
%

\subsection{Encoder}
The encoder aims to encode a post to a vector representation.
Given a post $\bm{x}=x_1x_2\cdots x_n$, the hidden states of the post $\bm{h}=(h_1,h_2, \cdot\cdot\cdot, h_n$) are obtained by a gated recurrent unit (GRU) ~\cite{chung2014empirical}, as follows:
\begin{equation}
    \label{eq:encoder}
    h_t = \mathbf{GRU}(h_{t-1}, x_t)
\end{equation}
where $x_t$ is the embedding of the $t$-th word.

\subsection{Profile Detector}
\label{sec:profile_detector}
The profile detector has two roles: first to detect whether the post should be responded with the agent profile,
and second to select a specific profile $<key,value>$ to be addressed in the decoder. 
The first role of the profile detector is defined by the probability $\bm{P}(z|\bm{x})$ ($z\in \{0,1\}$) where $z=1$ means the agent profile should be used.
For instance, if the post is \textit{``how old is your father''}, $\bm{P}(z=1|\bm{x}) \approx 0$, while if the post is \textit{``how old are you''},
$\bm{P}(z=1|\bm{x}) \approx 1$. 

$\bm{P}(z|\bm{x})$ is a binary classifier trained on supervised data. 
More formally, the probability is computed as follows:
\begin{equation}
    \label{eq:profile_detector}
    \bm{P}(z|\bm{x}) = \bm{P}(z|\widetilde{h}) 
    = \sigma(W_p \widetilde{h})
\end{equation}
where $W_p$ is the parameter of the classifier and $\widetilde{h}=\sum_j h_j$, simply the sum of all hidden states, but other elaborated methods such as attention-based models are also applicable.

The {\bf second role} of the profile detector is to decide which profile value should be addressed in a generated response.
This is implemented as follows:
\begin{equation}
    \label{eq:profile_attn}
    \begin{split}
        \beta_i & = \mathbf{MLP}([\widetilde{h},k_i,v_i]) \\
        & = f(W \cdot [\widetilde{h}; k_i;v_i])
    \end{split}
\end{equation}
where $W$ is the weight and $k_i/v_i$ is the embedding of a profile key/value respectively, they are all parameters of our model. $\widetilde{h}=\sum_j h_j$ is the representation of the post. $f$ is a nonlinear activation function, in this equation $f$ is a $softmax$ function over all $\beta_i$.
The above equation can be viewed as a multi-class classifier that produces a probability distribution over profile keys. 

The optimal profile value is selected with the maximal probability: $\widetilde{v}= v_j ~~~where~ j=argmax_i(\beta_i)$.
As long as a profile value $\widetilde{v}$ is obtained,
the decoding process will be determined by the bidirectional decoder, as follows:
\begin{equation}
    \label{eq:profile_seq}
    \begin{split}
    & \bm{P}^{bi}(\bm{y}|\bm{x},\{<k_i,v_i>\}) = \bm{P}^{bi}(\bm{y}|\bm{x},\widetilde{v}) \\
    \end{split}
\end{equation}
%

\subsection{The Bidirectional Decoder}
This decoder aims to generate a response in which a profile value will be mentioned. Inspired by \cite{mou2016sequenceBackwardForward}, we design a bidirectional decoder which consists of a backward decoder and forward decoder, but with a key difference that a position detector is employed to predict a start decoding position. 

Suppose a generated response is $\bm{y}$ = ($\bm{y^b},\widetilde{v},\bm{y^f}$) = ($y^b_1$, $\cdots$, $y^b_{t-1}$, $\widetilde{v}$, $y^f_{t+1}$, $\cdots$, $y^f_{m}$) where $\widetilde{v}$ is a selected profile value. The bidirectional decoder will generate $\bm{y^b}$ in a backward direction and $\bm{y^f}$ forward.
The backward decoder ($\bm{P^b}$) generates $\bm{y^b}$ from the given profile value $\widetilde{v}$ to the start of the response. The forward decoder ($\bm{P^f}$)\footnote{Note that this decoder is different from $\bm{P}^{fr}(y_t|y_{<t},\bm{x})$.} generates $\bm{y^f}$ from $\widetilde{v}$ to the end of the response, but takes as input the already generated first half, $\bm{y^b}$. The process is defined formally as follows:
\begin{equation}
    \label{eq:decoder}
    \begin{split}
        & \bm{P}^{bi}(\bm{y}|\bm{x},\widetilde{v}) = \bm{P}^{b}(\bm{y^b}|\bm{x},\widetilde{v})*\bm{P}^{f}(\bm{y^f}|\bm{y^b},\bm{x},\widetilde{v})) \\
        & \bm{P}^{b}(\bm{y^b}|\bm{x},\widetilde{v})=\prod_{j=t-1}^1 \bm{P}^{b}(y^b_{j}|y^b_{>j},\bm{x},\widetilde{v}) \\
        & \bm{P}^{f}(\bm{y^f}|\bm{y^b},\bm{x},\widetilde{v})=\prod_{j=t+1}^m \bm{P}^{f}(y^f_{j}|y^f_{<j},\bm{y^b},\bm{x},\widetilde{v})
    \end{split}
\end{equation}

In order to encode more contexts in the forward decoder, the first half of generated response ($\bm{y^b}$), along with the profile value ($\widetilde{v}$), serves as initial input to the forward decoder.
%
%
The probability $\bm{P}^{b}$ and $\bm{P}^{f}$ is calculated via
\begin{equation}
    \label{eq:decoder_output}
    \begin{split}
    & \bm{P}^{b}(y^b_{j}|y^b_{>j},\bm{x},\widetilde{v}) 
    \propto \mathbf{MLP}([s^b_j;y^b_{j+1};c^b_j]) \\
    & \bm{P}^{f}(y^f_{j}|y^f_{<j},\bm{y^b},\bm{x},\widetilde{v})
    \propto \mathbf{MLP}([s^f_j;y^f_{j-1};c^f_j])    
    \end{split}
\end{equation}
where $s^{(*)}_j$ is the state of the corresponding decoder, $c^{(*)}_j$ is the context vector, and $* \in \{b,f\}$ where $b$ indicates the backward decoder
and $f$ the forward.
The vectors are updated as follows:

\begin{equation}
    \begin{split}
    \label{eq:decoder_details}
    & s^{(*)}_j = \mathbf{GRU}(s^{(*)}_{j+l},[y^{(*)}_{j+l};c^{(*)}_j]) \\
    & c^{(*)}_j = \sum_{t=1}^{n} \alpha^{(*)}_{j,t}h_t 
    \end{split}
\end{equation}
where $\alpha^{(*)}_{j,t} \propto \mathbf{MLP}([s^{(*)}_{j+l},h_t])$ can be viewed as the similarity between decoder state $s^{(*)}_{j+l}$ and encoder hidden state $h_t$, 
$l=1$ when $*=b$~(backward), and $l=-1$ when $*=f$~(forward). 
And these $\mathbf{MLP}$s have the same form as Eq.\ref{eq:profile_attn}, but with different parameters.

\subsection{Position Detector}
The position detector is designed to provide more supervision to the bidirectional decoder, which is only used during training. 
As mentioned, the bidirectional decoder starts from a profile value to generate the entire sequence at the test stage. However, in our training dataset, the profile values may be rarely mentioned in the responses.
For instance, given the profile key value pair $<$爱好, 冰球$>$ ($<hobby, hockey>$), the value {\it 冰球(hockey)} rarely occurs in the training corpus.
In other words, even though we have a training instance $(\bm{x},\bm{y},<k,v>)$, the value ($v$) may not occur in $\bm{y}$ at all.
Hence, the bidirectional decoder is not aware from which word decoding should start.
This leads to the discrepancy between training and test: during training, the decoder is unaware of the start decoding position but during test, the start decoding word is given. 

This issue makes our work differ substantially from previous approaches
where supervision is directly observable either between post and response \cite{gu2016CopySeq2Seq} or between response and knowledge base~\cite{yin2015GenQA}. Experiments also show that the position detector contributes much to the performance improvement than a random position picking strategy \cite{mou2016sequenceBackwardForward}.

The position detector is designed to provide a start decoding position to the decoder during training. For instance,
given a post $\bm{x}=$``你-1 有-2 什么-3 特长-4 ？-5 (what's your speciality?)\footnote{The number indicates the position of each word.}'' and a response $\bm{y}=$``我-1 非常-2 擅长-3 小提琴-4(I am good at playing violin)'',
and a profile key value pair ``$<$特长, 钢琴$>$ ($<hobby, piano>$)'', the position detector will predict that ``小提琴-4 (violin)'' in the response can be replaced 
by the profile value ``钢琴(piano)'' to ensure grammaticality. The predicted position ``小提琴-4 (violin)'' is then passed to the decoder (see Eq. \ref{eq:decoder}) to signal the start decoding position. 

In order to find an appropriate position at which the profile value can be replaced, we need to estimate the probability: $\bm{P}(j|y_1y_2\cdots
y_m,<k,v>)),1 \leq j \leq m$ which
indicates how likely the word $y_j$ can be replaced by the profile value $v$.

We apply a simple technique to approximate the probability: a word can be replaced by a given profile value if the word has maximal similarity.
\begin{equation}
    \label{eq:position_vector}
    \bm{P}(j|\bm{y},<k,v>)) \propto cos(y_j, v)
\end{equation}
where $cos(y_j,v)$ denotes the cosine similarity between a word in a response and a profile value. More elaborated techniques, for instance, language models, will be studied as future work.
    

\subsection{Loss Function and Training}
Two loss functions are defined: one on the generation probability and the other on the profile detector.
%
%
The first loss is defined as below:
\begin{equation}
    \begin{split}
    \label{eq:loss_decoder}
    & \mathcal{L}_1(\theta,D^{(c)},D^{(x,y)}) \\
    = & -\sum_{(\bm{x},\bm{y}) \in D^{(c)} \cup D^{(pr)}} log \bm{P}(\bm{y}|\bm{x},\{<k_i,v_i>\})  \\
    = & -\sum_{(\bm{x},\bm{y}) \in D^{(c)}}log \bm{P}^{fr}(\bm{y}|\bm{x}) \\ 
    & -\sum_{(\bm{x},\bm{y}) \in D^{(pr)}}log \bm{P}^{bi}(\bm{y}|\bm{x},\widetilde{v})
    \end{split}
\end{equation}
The first term is the negative log likelihood of observing $D^{(c)}$ and the second term for $D^{(pr)}$.
$\widetilde{v}$ is a word in $\bm{y}$ whose position is predicted by the position detector during training.
$D^{(pr)}$ consists of pairs where a post is related to a profile key and its response gives a meaningful reaction to the post, and $D^{(c)}$ has only general post-response pairs.

The two decoders ($\bm{P}^{fr}$ and $\bm{P}^{bi}$) have no shared parameters.
Since the number of instances in $D^{(c)}$ is much larger than that of $D^{(pr)}$, we apply a two-stage training strategy:  $D^{(c)}$ will be used to train  $\bm{P}^{bi}$ at the early stage for several epoches, where $\widetilde{v}$ is a randomly chosen word in a response, and then $D^{(pr)}$ for further training at the later stage.


The above formulation generally adopts the hard form of $\bm{P}(z|\bm{x})$ (see Eq. \ref{eq:profile_detector}): $\bm{P}(z=1|\bm{x})=1$ for profile-related pairs and $\bm{P}(z=1|\bm{x})=0$ for others. In order to better supervise the learning of the profile detector, we define the second loss and add it to the first one with a weight $\alpha$ as the overall loss (i.e., $\mathcal{L}=\mathcal{L}_1+\alpha\mathcal{L}_2$):
\begin{equation}
\label{eq:loss_profile}
    \begin{split}
    & \mathcal{L}_2(\theta, D^{(pb)}, D^{(pr)}) \\
    = & - \sum_{(\bm{x},\bm{y},z) \in D^{(pb)}} log \bm{P}(z|\bm{x}) \\
    & - \sum_{(\bm{x},\bm{y},\widehat{k}) \in D^{(pr)}} \sum_{j=1}^K \widehat{\beta_j} log \beta_j
    \end{split}
\end{equation}
where the first term is for binary prediction of using profile or not, and the second for profile key selection. $\widehat{k}$ is the profile key whose value should be addressed, $K$ is the total number of keys, $\bm{\beta}$ is the predicted distribution over profile keys as defined by Eq.~\ref{eq:profile_attn}, and $\bm{\widehat{\beta}}$ is one-hot representation of the gold distribution over keys. $<\bm{x},\bm{y},z>$ is obtained by manual annotation while $(\bm{x},\bm{y},\widehat{k})$ is obtained by matching the keywords and synonyms in the profile with the post, which is noisy. This works well in practice and reduces manual labors largely.

\section{Experiment}
\subsection{Data Preparation}
We prepare several datasets:
\\
\textbf{Weibo Dataset (WD) - $D$}: 
We collect $9,697,651$ post-response pairs from Weibo.
The dataset is used for training $\bm{P}^{fr}(\bm{y|x})$ and $\bm{P}^{bi}(\bm{y|x},\widetilde{v})$ at the early stage and 7,000 pairs are used for validation to make early stop.
\\
\textbf{Profile Binary Subset (PB - $D^{(pb)} \in D$)}: 
We extract $76,930$ pairs from WD for 6 profile keys (\{{\it name, gender, age, city, weight, constellation}\}) with about 200 regular expression patterns. 
The dataset is annotated by 13 annotators. Each pair is manually labeled to {\it positive} if a post is asking for a profile value and the response is a logic reaction to the post, or {\it negative} otherwise.

This dataset is used to train the binary classifier ($\bm{P}(z|\bm{x})$)~(see $D^{(pb)}$ in Eq. \ref{eq:loss_profile}). 3,000 pairs are used for test and the remainder for training.
The statistics of the dataset is shown in the supplementary file.
\\
\textbf{Profile Related Subset (PR - $D^{(pr)} \in D^{(pb)}$)}:
This dataset only contains pairs whose posts are positive in PB. In total, we have $42,193$ such pairs.
This dataset is used to train the bidirectional decoder.
\\
\textbf{Manual Dataset (MD)}:
This dataset has 600 posts written by 4 human curators, including 50 negative and 50 positive posts for each key. A positive post for a profile key (e.g., {\it how old are you?}) means that it should be responded by a profile value, while a negative post (e.g., {\it how old is your sister?}) should not.
This dataset is used to test the performance on real conversation data rather than social media data.

All datasets are available upon request. {\bf Implementation details of the model are shown in the supplementary.}

\subsection{Human Evaluation}
We evaluate our model at both post and session level. At the post level, we define three metrics ({\it naturalness, logic, and correctness}) to evaluate the response generated by each model. At the session level, we evaluate the models from the aspects of {\it consistency and variety} to justify the performance in the real conversational setting.

We name our model \textit{Identity-Coherent Conversation Machine (ICCM)} and compare it with several baselines:
\\
\textbf{Seq2Seq}: a general sequence to sequence model~\cite{sutskever2014seq2seq}.\\ 
\textbf{Seq2Seq + Profile Value (+PV)}: if the profile detector decides that a profile value should be used ($\bm{P}(z|\bm{x}) > 0.5$), the response is simply the value of the key decided by the profile detector (see Eq. \ref{eq:profile_attn}); otherwise, a general seq2seq decoder will be used.   \\
\textbf{Seq2Seq + Profile Value Decoding (+PVD)}: the response is generated by a general seq2seq decoder which starts decoding forwardly from the value of the selected key.
\\
\textbf{ICCM-Pos}: Instead of using a predicted position obtained by the position detector to start the decoding process, the bidirectional decoder in this setting randomly picks a word in a response during training, the same as \cite{mou2016sequenceBackwardForward}. 

\begin{table*}
\vspace{3mm}
\newcommand{\tabincell}[2]{\begin{tabular}{@{}#1@{}}#2\end{tabular}}
\centering{
	\begin{tabular}{|l|l l|}
	    \hline
	    & \textbf{Chinese} & \textbf{English (Translated)} \\
	    \hline
         & Post:你是不是美少女呀？ & Post: Are you a beautiful girl ? \\
        \cline{2-3}
        Seq2Seq & 是呀。 & Yes, I am. \\
        Seq2Seq +PV & \underline{男生} & \underline{Boy} \\
        Seq2Seq +PVD & \underline{男生}。 & \underline{Boy}. \\
        ICCM-Pos (ours) & 你是\underline{男生}？ & Are you a \underline{Boy}? \\
        ICCM (ours) & 我是\underline{男生}！ & I am a  \underline{Boy}! \\
	    \hline
	\end{tabular}}
	\caption{Sample responses generated by our model and baselines. The profile value in the agent profile is marked in \underline{underline}.}
	\label{table_case_post}
\end{table*}

\subsubsection{Post-level Evaluation}
To conduct post-level evaluation, we use 600 posts from MD, 50 positive/negative posts respectively for each key. Each post is input to all the models to get the corresponding responses. Thus, each post has 5 responses and these responses are randomly shuffled and then presented to two curators. Post-response pairs are annotated according to the following metrics, based on a 1/0 scoring schema:
\\
\textbf{Naturalness (Nat.)} measures the fluency and grammaticality of a response. Too short responses will be judged as lack of naturalness.   
\\ 
\textbf{Logic} measures whether the response is a logical reaction to a post. For instance, for post ``how old are you'', a logical response could be ``I am 3 years old'' or ``I do not know''. \\
\textbf{Correctness (Cor.)} measures whether the response provides a correct answer to a post given the profile. For instance, for post ``how old are you'', if the profile has a key value pair like $<age,3>$, responses like ``I am 18'' will be judged as wrong. 

Each response is judged by two curators. The Cohen's Kappa statistics are 0.46, 0.75 and 0.82 for naturalness, logic, and correctness respectively. Naturalness has a rather lower Kapp because it is more difficult to judge.

\begin{table} [!htp]
\vspace{1mm}
\centerline{
\begin{tabular}{c|c|c|c}
    \hline
    Method & Nat. & Logic & Cor. \\
    \hline
    Seq2Seq &       71.4\% & 38.7\% & 22.3\% \\
    Seq2Seq +PV &  85.4\% & 51.3\% & 40.2\% \\
    Seq2Seq +PVD &   84.7\% & 51.1\% & 40.3\% \\
    ICCM-Pos (ours) & 87.4\% & 50.0\% & 41.8\% \\
    ICCM (ours) &     \textbf{88.9\%} & \textbf{55.9\%} & \textbf{44.2\%} \\
    \hline
\end{tabular}}
\caption{Evaluation of responses to the 600 posts from MD.}
\label{table_manually}
\end{table}

Results in Table \ref{table_manually} support the following statements: {\bf First}, our model is better than all other baselines in all metrics, indicating that our model can generate more natural, logical, and correct responses; {\bf Second}, in comparison to simply responding with a profile value (Seq2Seq+PV) where the responses are generally too short, our model can generate more natural responses; {\bf Third}, the position detection contributes to better generation, in comparison to a random position (ICCM vs. ICCM-Pos). 
Exemplar responses generated by these models are shown in Table \ref{table_case_post} which also demonstrate the effectiveness of our model.

\subsubsection{Session-level Evaluation}
In order to compare these models in real conversation sessions, we randomly generate sessions based on MD. For each profile key, we randomly choose 3 {\it  positive} posts\footnote{A \textit{positive} post must be responded with a profile value.} from MD, generate responses to the 3 posts for each model, and obtain a session of 3 post-response pairs. In this way, 240 sessions are generated, and each key has 40 sessions. The sessions are manually checked with the following metrics:
\\
\textbf{Consistency} measures whether there are contradictory responses with respect to the given profile. Score 1 indicates that all the three responses are consistent to the profile, and score 0 otherwise. \\
\textbf{Variety} measures the language variety of the three responses in a session. Score 1 indicates that the linguistic patterns and wordings are different between any two of them, and score 0 otherwise. 

Results are shown in Table \ref{table_manually_session} and we show some session examples in Table \ref{session-level-examples}.

\begin{table} [!htp]
	\begin{tabular}{l|l}
		\hline
		\textbf{Chinese} & \textbf{English(Translated)} \\
		\hline
		U:你还没说你几岁呢 & U:You haven't told me \\
		& your age. \\
		S:我三岁了 & S:I'm three years old. \\
		U:你今年有15了不 & U:Are you 15 years old \\
		& or not?\\
		S:我还没到呢 & S:I'm not yet. \\
		U:你多大啦 & U:How old are you?\\
		S:3岁了 & S:Three years old. \\
		\hline 
	\end{tabular} 
	\caption{Samples of consistent conversations generated by our model. $U/S$ indicates User/System.}
	\label{session-level-examples}
\end{table}

We can clearly see the following observations: \\
\textbf{1)} Our model is remarkably better than all the baselines w.r.t both metrics. Results of our model against Seq2Seq+PVD indicate that the bidirectional decoder can generate responses of much richer language variety. The results of ICCM-Pos show that the position detector improves consistency and variety remarkably.
\\
\textbf{2)} if simply respond with a profile value (Seq2Seq+PV), the model can obtain good consistency but very bad language variety, which is in line with the intuition.
\\
\textbf{3)} The general Seq2Seq model is too weak to generate consistent or linguistically various responses.

\begin{table} [!htp]
\vspace{1mm}
\centerline{
\begin{tabular}{c|c|c}
    \hline
    Method & Consistency & Variety \\
    \hline
    Seq2Seq &       2.1\% & 1.6\% \\
    Seq2Seq +PV &   58.3\% & 2.1\% \\
    Seq2Seq +PVD &  47.5\% & 10.0\% \\
    ICCM-Pos (ours) & 46.7\% & 21.2\% \\
    ICCM (ours) &     \textbf{60.8\%} & \textbf{33.3\%} \\
    \hline
\end{tabular}}
\caption{Consistency and variety on the 240 sessions generated from MD.}
\label{table_manually_session}
\end{table}

\subsection{Automatic Evaluation}
We also present results of automatic evaluation for the profile and position detector.
\subsubsection{Profile Detection}
The profile detector is evaluated from two aspects:
whether a profile should be used or not ($\bm{P}(z=1|\bm{x})$), and whether a profile key is correctly chosen.
Note that the prediction of profile key selection is cascaded on that of $\bm{P}(z=1|\bm{x})$.

\begin{table} [!htp]
\vspace{1mm}
\centerline{
\begin{tabular}{c|c|c}
    \hline
    Dataset (\# samples) & Binary profile & Key selection \\
    \hline
    PB (3000) & 85.1\% & 74.8\% \\
    \hline
    MD (600) & 82.0\% & 70.5\% \\
    \hline
\end{tabular}}
\caption{Classification accuracy of the profile detector.}
\label{table_profile_1}
\end{table}


The classifiers are trained on Weibo social data. Results in Table \ref{table_profile_1} show that the profile detector obtains fairly good accuracy.
But the classifiers have a remarkable drop when test on the manual dataset (comparing two rows: MD(600) vs. PB(3000)). This indicates the difference between Weibo social data and real human conversations.



\subsubsection{Position Detection}
As mentioned previously, the position detector plays a key role in improving the naturalness, logic, and correctness of responses (see ICCM vs. ICCM-pos in Table \ref{table_manually}), and the consistency and variety of conversational sessions (see Table \ref{table_manually_session}). Thus, it is necessary to evaluate the performance of this module separately.

\begin{table} [!htp]
\vspace{1mm}
\centerline{
\begin{tabular}{c|c|c|c}
    \hline
    Profile Key & Acc & Profile Key & Acc \\
    \hline
    Name    & 35.0\% & Gender        & 96.0\% \\ 
    Age     & 98.5\% & Weight        & 85.5\% \\
    City    & 99.0\% & Constellation & 100.0\% \\
    \hline
\end{tabular}}
\caption{Accuracy for predicting the start decoding position.}
\label{table_position}
\end{table}

We randomly sample 200 post-response pairs from PR for each key (1200 pairs in total), and then manually annotate the optimal position from which decoding should start. The results are shown in Table \ref{table_position}. The position for most keys can be estimated accurately while for {\it name} the prediction is bad. This is because the value of the key rarely occurs in our corpus, and the embeddings of such values are not fully trained.
Nevertheless, the results are better than a random word picking strategy (ICCM vs. ICCM-Pos). 

\subsection{Extensibility}
The effectiveness of our model is verified on six profile keys, but much manual labors are required. We will show the extensibility of the model by evaluating it on four additional keys: {\it hobby, idol, speciality, and employer}.


Firstly, for the 4 keys,
we extract $16,332$ post-response pairs from WD with $79$ hand-crafted patterns and each pair is noisily mapped to one of the keys with these patterns. These new pairs, along with the old pairs on the six pairs, are used to retrain the model. Secondly, we construct a test dataset consisting of 400 posts, 50 positive and 50 negative human-written posts for each key. Responses from our model and Seq2Seq are obtained and then evaluated. The manual labor exists only in hand-crafting the 79 patterns.

Results show that our model has a relative 10\% drop on the new keys with respect to {\it logic and correctness}, and remains unchanged in {\it naturalness}. Nevertheless, our model is still much better than the Seq2Seq model. The baseline has no drop in {\it naturalness and logic} because this model does not rely on profile.


\begin{table} [!htp]
\vspace{1mm}
\centerline{
\begin{tabular}{c|c|c|c|c}
    \hline
    Dataset & Method & Nat. & Logic & Cor. \\
    \hline
    \multirow{2}{*}{6 keys} & Seq2Seq & 71.4\% & 38.7\% & 22.3\% \\
    & ICCM  & 88.9\% & 55.9\% & 44.2\% \\
    \hline
    \multirow{2}{*}{4 keys} & Seq2Seq & 75.6\% & 39.9\% & 17.9\% \\
    & ICCM  & 88.8\% & 50.9\% & 39.6\% \\
    \hline
\end{tabular}}
\caption{Extensibility evaluation on 4 new keys.}
\label{table_compare}
\end{table}

\section{Conclusion and Future Work}
We present a model that can generate responses that are coherent to a pre-specified agent profile. 
Instead of learning personality from dialogue data, our work can assign a desired identity to a chatbot.
Experiments show that our model is effective to generate more coherent and various conversations. 



Our work is a very small step to endow a chatbot with its own personality, which is an important issue for a chatbot to pass the Turing Test. 
There are many future directions:
\\
\textbf{Conversation style}: we have demonstrated that general conversation data can be used to generate profile-coherent responses. Can conversation style that is coherent to a chatbot's personality be modeled without stylistic dialogue data? Learning conversation styles such as talking like young girls or old adults, and  introverts or extroverts,  will be an interesting direction. 
\\
\textbf{Semantic reasoning}: Endowing a chatbot with personality/identity arises the issue of semantic reasoning. When the user asks ``are you married?'' or ``do you play women basketball?'' to a five-year-old boy agent, a coherent response requires common sense knowledge and reasoning, however, this is extremely challenging.

\bibliography{emnlp2017}
\bibliographystyle{emnlp_natbib}

\appendix

\section{Statistics for Profile Binary Dataset}
\label{appendix:statistics}

We extract $76,930$ pairs from WD for 6 profile keys (\{{\it name, gender, age, city, weight, constellation}\}) with about 200 hand-crafted patterns. And then they are annotated to {\it positive} or {\it negative}.
A {\it positive} post asks for a profile key of chatbot and its response gives a meaningful reaction, such as {\it ``Could you tell me your age''}, while a {\it negative} post is irrelevant to any profile key, such as {\it ``Guess how old I am''}. 
The statistics of the dataset is shown in Table \ref{table_profile_distribution}. 

\begin{table} [!htp]
\vspace{1mm}
\centerline{
\begin{tabular}{|c|c|c|}
    \hline
    Profile Key & Positive & Negative \\
    \hline
    Name & 6,966 & 3,442 \\
    \hline
    Gender & 7,665 & 8,259 \\
    \hline
    Age & 6,038 & 3,309 \\
    \hline
    City & 6,264 & 8,350 \\
    \hline
    Weight & 6,856 & 3,800 \\
    \hline
    Constellation & 8,404 & 7,577 \\
    \hline
\end{tabular}}
\caption{Statistics of the profile binary dataset. The data is mined from Weibo Dataset by hand-crafted patterns and manually labeled.}
\label{table_profile_distribution}
\end{table}

\section{Implementation Details}
\label{appendix:imple-details}

In our experiments, the encoder and attentive decoders are all have 4 layers of GRUs with a $512$-dimensional hidden state. The dimension of word embedding is set to $100$. The vocabulary size is limited to $40,000$. 
The word embeddings are pre-trained on an unlabeled corpus (about $60,000,000$ Weibo pairs) by word2vec. And the other parameters are initialized by sampling from a uniform distribution $U(-sqrt(3/n), sqrt(3/n))$, where $n$ is the dimension of parameters.
Training is conducted by stochastic gradient descent (SGD) with a mini-batch of 128 pairs.
The learning rate is initialized with $0.5$ and the decay factor is $0.99$.

\end{CJK*}
\end{document}